\documentclass[runningheads]{llncs}
\pdfoutput=1
\usepackage{graphicx}
\usepackage{color}
\usepackage{adjustbox}
\usepackage{tabularx}

\newcommand*\samethanks[1][\value{footnote}]{\footnotemark[#1]}

\begin{document}
\title{Video-Based Convolutional Attention \\for Person Re-Identification}
\author{Marco Zamprogno\thanks{Indicates equal contribution.} \and
Marco Passon\samethanks[1] \and
Niki Martinel \and
Giuseppe Serra \and
Giuseppe Lancioni \and
Christian Micheloni \and
Carlo Tasso \and
Gian Luca Foresti
}
\authorrunning{M. Zamprogno et al.}
% Marco Zamprogno, Marco Passon, Niki Martinel, Giuseppe Lancioni, Christian Micheloni, Giuseppe Serra, Carlo Tasso, Gian Luca Foresti
% University of Udine
\institute{Universit\`a degli Studi di Udine, Udine (UD), Italy}
\maketitle              % typeset the header of the contribution
\begin{abstract}
In this paper we consider the problem of video-based person re-identification, which is the task of associating videos of the same person captured by different and non-overlapping cameras.
We propose a Siamese framework in which video frames of the person to re-identify and of the candidate one are processed by two identical networks which produce a similarity score.
We introduce an attention mechanisms to capture the relevant information both at frame level (spatial information) and at video level (temporal information given by the importance of a specific frame within the sequence).
One of the novelties of our approach is given by a joint concurrent processing of both frame and video levels, providing in such a way a very simple architecture. Despite this fact, out approach achieves better performance than the state-of-the-art on the challenging iLIDS-VID dataset.
% Despite being a simple architecture, our approach achieves better performance than the state-of-the-art on the challenging iLIDS-VID dataset.

\end{abstract}
\section{Introduction}
Given an image or video of a person taken from one camera, the Re-Identification task (ReID) is the process of re-associating the person by analyzing images or videos taken from a different camera with non-overlapping field of view. Although humans can easily re-identify others by leveraging descriptors based on the person's face, height, clothing, and walking pattern, ReID is a difficult problem for a machine to solve, since it should deal with features between cameras like different lighting conditions, different point of views or person occluded by objects or other people.

Traditionally many attempts to explore the problem has been proposed for still images (\textit{e.g.},~\cite{Martinel2012c,Martinel2017,Lisanti2017,Martinel2018,Martinel2018a,Lisanti2019}), while recently some research groups have experimented approaches based on video images~(\textit{e.g.},~\cite{DBLP:journals/corr/ZhengYH16}).
Using videos for Re-Identification provides several advantages over still images. The video setting is a more natural way to perform Re-Identification, as a person will normally be captured by a video camera producing a sequence of images rather than a single still image. Given the availability of sequences of images, temporal information related to a person motion may help to disambiguate difficult cases that arise when trying to recognize a person in a different camera. Furthermore, sequences of images provide a larger number of samples of a person appearance, thus allowing a better appearance model to be built. On the other hand, this large set of information needs to be treated properly.  %The existence of a large number of samples makes it easier to train machine learning algorithms, and neural networks in particular.

To address this challenge, in this paper we propose an approach to the problem of video-based person re-identification that is characterized by two main aspects.
First, we propose a deep neural network architecture based on a Siamese framework \cite{mclaughlin2016recurrent} which evaluates the similarity of the query video to a candidate one.
%The reason to choose Siamese models is to overcome the limited amount of training samples some identities have\cite{yi2014deep}.
Second, we introduce a novel spatio-temporal attention mechanism with the aim to select relevant information from different areas of the frames of the input video, and from their evolution over time. 
Attention mechanisms have been largely exploited in a variety of different implementations and in many different domains of Deep Learning such as Natural Language Processing \cite{passon2019keyphrase} and Computer Vision \cite{song2017end}. The intuition behind Attention in Computer Vision is to mimic the human visual process. Humans give different importance to different areas in an image as they are able to focus on 'hot' areas and neglect others~\cite{7900174}. 
%Attention introduces sets of weights that gives different values to these different areas. 
This improves greatly the ability to recognize structures and patterns in otherwise flat data. Nevertheless there are relatively few attempts to use Attention in the field of Automatic Re-Identification.
\cite{liu2017end} proposes integrating a soft attention based model in a Siamese network to  focus adaptively on the important local regions of an input image pair.
\cite{xu2017jointly} uses a spatial pyramid layer as the component attentive spatial pooling to select important regions in spatial dimension.
\cite{song2017end} proposes a spatial attention module focused on recognizing the skeleton to identify the poses, and then a temporal module to recognize the actions.

Unlike other approaches, which use at least two separate modules to identify spatial and temporal features, we use a joint module to identify both at the same time. This allows us to define a simpler architecture which provides state-of-the-art performance on the well-known iLIDS-VID dataset. 

\section{Related Work}
The interest for video-based Person Re-Identification has increased significantly in recent years \cite{Vezzani:2013:PRS:2543581.2543596}.
The aim of the first works was to manually extract feature representations invariant to changes in poses, lighting conditions, and viewpoints. Using these features, they proposed distance metrics to measure the similarity between two images. In particular, one of the first studies  computes the spatio-temporal stable region with foreground segmentation \cite{gheissari2006person}; while \cite{cong2009video} employs more compact spatial descriptors and color features, constructed by using the manifold geometry structure in video sequences.

With the advent of Deep Learning approaches, Convolutional Neural Networks (CNNs) have been introduced in visual recognition tasks yielding to considerable improvements in the performance \cite{krizhevsky2012imagenet} with respect to more classical solutions~\cite{Rani2015}. In fact, CNNs are able to extract different features from a given image, representing them as a set of output maps avoiding manual effort in feature engineering. 
Image-based Automatic Person Re-Identification is one of the fields in which CNNs achieved remarkable results \cite{qian2017multi,ustinova2017multi,varior2016siamese,xiao2016learning,zhang2016learning,hadsell2006dimensionality}. %In these works, the feature vectors are extracted from images using deep network architectures, and these feature vectors are successively employed to perform Re-Identification.

However, considering that Person Re-Identification is usually done in settings that involve, for example, surveillance cameras, it is easy to argue that image-based person re-identification is no more an adequate schema to address current needs.

This led to most recent works that began exploring video-based person re-identification \cite{li2017video,liu2015spatio,mclaughlin2016recurrent,wang2014person,xu2017jointly,yan2016person,zhou2017see,zhu2018video}, a setting closer to real-world applications. Videos have the advantage to contain temporal information that is potentially helpful in differentiating between persons. For example, in \cite{mclaughlin2016recurrent}, the proposed CNN model extracts features from subsequent video frames that are fed through a recurrent final layer in order to combine frame-level features and video-level features.

Not all the parts of an image or of a video are equally important and humans place more focus only on some of them, assigning little to no importance to the rest. This attention mechanism has been adopted in a variety of applications, such as machine translation \cite{bahdanau2014neural}, action recognition \cite{sharma2015action}, image recognition \cite{ba2014multiple} and caption generation \cite{xu2015show}. Recently, Attention models \cite{sharma2015action,song2017end} have been proposed for video and image understanding. These models assign weights to different parts of each frame, making some of them more important than others. In particular, \cite{liu2017end} proposes integrating a spatial attention based model in a siamese network to adaptively focus on the important local parts of an input image pair. 
%\cite{zhou2017see} uses a Recurrent Neural Network (RNN) to generate temporal attentions over frames, so that the model can focus on the most discriminative ones in a video for Person Re-Identification.

With respect to the existing literature,~\cite{zhou2017see} and~\cite{rao2018videobased} are the most similar to our approach. \cite{zhou2017see} uses a Recurrent Neural Network (RNN) to generate temporal attentions over frames so that the model can focus on the most discriminative ones in a video. \cite{rao2018videobased} instead directly calculates the attention scores on frame-based features, using a simple architecture with two separate temporal and spatial modules.
Our approach exploits a single attentive module to extract both temporal and spatial features from frames at the same time, resulting in an even simpler architecture that provides state-of-the-art performance.

\section{The Proposed Approach}

%\begin{figure}[t]
%\centering
%\makebox[\textwidth][c]{\includegraphics[width=1\textwidth]{imgs/SchemaAttentivo.png}}%
%\caption{Attentive Convolutive LSTM Network scheme. The attentive map is refined every step %t.\label{fig:attentivo}}
%\end{figure}

\begin{figure}[hbt]
\centering
\includegraphics[width=\textwidth, keepaspectratio]{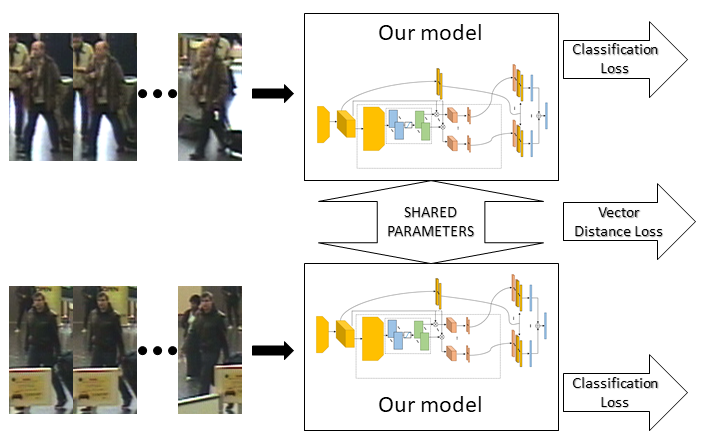}
\caption{Siamese network scheme. Each network receives as input a person image sequence to classify.
The loss is calculated as the sum of the classification error of each network, plus the Euclidean distance between the two descriptive vectors, which should be close to zero if the two sequences belong to the same person, or close to one if they belong to different people.\label{fig:siamese}}
\end{figure}

The proposed approach (see Fig. \ref{fig:siamese}) is based on a Siamese network \cite{mclaughlin2016recurrent}. This schema is composed by two identical networks, or branches, in which the first is fed with the query video and the second with the candidate video to be compared.
Each branch includes a sequence of modules that will be described in details in next sub-sections. The parameters of the two branches are shared.
The output of the Siamese network is a value that represents the similarity of the two input video sequences in terms of the distance between their respective features vectors, which should be close to zero if they belong to the same person, close to one otherwise.

\subsection{Spatio-Temporal Attentive Module}
The Spatio-Temporal Attentive Module is the core module of the proposed architecture. %It is based on the saliency prediction model proposed in~\cite{7900174}. 
It aims to identify the portions of a frame which an human eye would normally focus on. Those areas should contain relevant spatial information, and we want to exploit them to improve the re-identification performance. Since the input frames are enhanced with the temporal information of the optical flow, both spatial and temporal features will be exploited by this network.

\begin{figure}[hbt]
\centering
\makebox[\textwidth][c]{\includegraphics[width=1\textwidth]{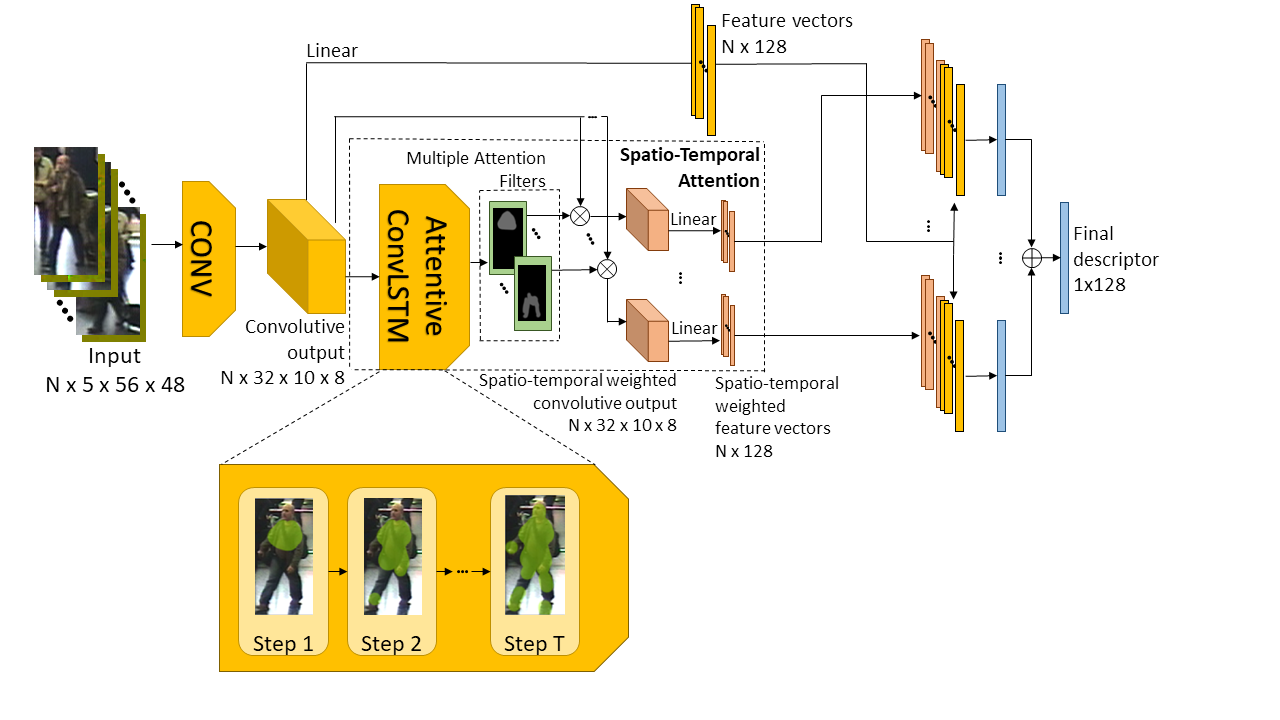}}%
\caption{Detailed Network scheme. The main blocks are the initial convolutional network, the Spatio-Temporal Attention Module, and the final part which performs averaging and normalization. The bottom part gives an idea of the multiple refining steps.\label{fig:rete}}
\end{figure}

Inspired by~\cite{7900174}, we propose to use a particular combination of convolutional network and LSTM, called Attentive ConvLSTM, capable of working on spatial features, in which the internal state of the network is given by the standard LSTM state equations where the matrix products between weights and inputs are replaced by convolutional operators. The ability to work with sequences is exploited to process input spatial features iteratively. The general idea of how this module works is shown in the bottom part of Fig.\ref{fig:rete}.

Our aim is to exploit attentive maps to better identify relevant features of frames and provide state-of-the-art performance while using a simple network. 
The architecture of each branch (see Fig. \ref{fig:rete}) is based on an initial convolutional network to reduce the image size, an attentive model to generate attentive maps, a fully connected layer to extract significant features from the original frames, and a final part where the features are combined.

More in details, the architecture consists of an ConvLSTM to recurrently processes attentive features at different locations of the frame, focusing on different regions of the tensor.
A stack of features $\mathbf{X}$ is repeatedly given in input to the LSTM, which sequentially updates an internal state based on three sigmoid activators. Update is performed by two blocks: the Attentive Model, and the ConvLSTM.
The Attentive Model generates an attention map using a convolutional layer that takes as input the original $\mathbf{X}$ and the previous hidden state, followed by a $\mathtt{tanh}$ activation function and another convolutional layer, and finally normalized with a $\mathtt{softmax}$ operator. The resulting output represents a normalized spatial attention map, which is then applied to the original $\mathbf{X}$ with an element-wise product, resulting in the filtered $\mathbf{X^\prime}$.
In ConvLSTM, each of the three sigmoid activators is given in input the sum of two different convolutive layers, the first taking as input $\mathbf{X^\prime}$ and the second taking as input the previous hidden state, and a bias. The output of the first sigmoid is then multiplied element-wise with the previous $\mathbf{X^\prime}$, the output of the second sigmoid is multiplied element-wise with the state of the LSTM memory cell, and the two resulting outputs are summed together and fed to a $\mathtt{tanh}$ activator. The result is multiplied element-wise with the output of the third sigmoid, and the resulting tensor is the new hidden state.

The Spatio-Temporal Attentive Module takes in input an image and produces in output multiple attentive maps, using an iterative refinement in $T$ steps (based on our preliminary experiments, we set $T=10$). We then apply those maps to the original input and obtain multiple different filtered outputs. Ideally, each filter should focus on a different spatio-temporal feature of the frame.

\subsection{Architecture Details}
The starting input (see Fig. \ref{fig:rete}) consists of video sequences composed by a batch of N frames, each frame has size $56\times 48$, with 3 channels for the YUV, plus 2 for the vertical and horizontal components of the optical flow, for a total of 5 channels.

The input is first processed through a convolutional network which consists of 3 stages, each composed by convolution, max-pooling, and nonlinear activations. Each convolution filter uses $5\times 5$ kernels with $1\times 1$ stride and $4\times 4$ zero padding. This outputs a batch of size $N\times 32\times 10\times 8$.

At this point, the model branches in two lines: the same input is passed to the Spatio-Temporal Attentive Module previously described, and to a fully connected layer preceded by a dropout applied with $p=0.6$ probability. The first aims to output multiple spatio-temporal-filtered feature vectors for each frame, and the second a general feature vector for each frame.

Spatio-Temporal Attention generates multiple attentive filters. Each of these filters has size $10\times 8$, is first normalized with a sigmoid between 0 and 1, and then applied with an element-wise multiplication to the original output of the first convolutional network, obtaining multiple blocks weighted with a different filter with the same dimension of the input, $N\times 32\times 10\times 8$; each of these blocks focus on a specific zone of the frames. A final fully connected layer generates, for each block, a batch of spatio-temporal-weighted feature vectors of size $N\times 128$. This final layer is also preceded by a dropout with p=0.6. In our model, since we generate 3 filters, we obtain 3 spatio-temporal-weighted feature vectors.

The two branches of the network are then merged together, and the general feature vectors are concatenated with each of the spatio-temporal weighted feature vectors, resulting in 3 combined-feature vectors of size $2N\times 128$. Finally, each of these batches is averaged, normalized using L2 normalization, and lastly summed together, obtaining a final feature descriptor of size $1\times 128$.

\section{Experimental Results}
%We first introduce the dataset used in our experiments and the experimental setup. In the last section we present the results of our experiments.
Our approach has been tested and evaluated on the public iLIDS-VID benchmark \cite{wang2014person}, since it is a challenging dataset that contains many occlusions, severe illumination changes and background clutters. It is also widely used in literature and it is then easier to fair compare our results. The iLIDS-VID dataset consists of videos of 300 distinct people. For each person there are two different video sequences, captured by two non-overlapping cameras. The video sequences have a varying number of frames, with the shortest sequence having 23 frames long and the longest having 192 frames, averaging at 73 frames.

\subsection{Experimental setup}\label{setup}
To be comparable with literature, we follow the experimental setup proposed by \cite{mclaughlin2016recurrent}. The dataset is randomly split in two: 50\% of the people form the training set and 50\% the test set. During the execution of the experiments, a different train/test split is computed for every repetition and the final results are then averaged. The network is trained for 1500 epochs using Stochastic Gradient Descent algorithm. One epoch consists in showing the Siamese network an equal number of positive sequence pairs and negative pairs, sampled randomly from all the persons in the training set, alternatively.

A positive sequence pair consists of two full sequences of arbitrary length, recorded by two different cameras, showing the same person. Analogously, a negative sequence pair shows two different persons. During the training phase, the length of the sequences is set to 16, that is, 16 consecutive frames belonging to a person are randomly sampled and used during this phase. As in \cite{wang2014person}, the first camera is the probe and the second the gallery.

All the images in the dataset go through a preprocessing step where they are converted from the RGB to the YUV color space and each color channel is normalized in order to have zero mean and unit variance. The three color channels are expanded with two more channels corresponding to the horizontal and vertical component of the optical flow computed between each pair of consecutive frames using Lucas-Kanade algorithm \cite{lucas1981iterative}. The two optical flow channels are normalized to bring them within the $[-1, 1]$ range.

Data augmentation is applied to each sequence during the training phase in order to increase the diversity of the training sequences. Each frame in the sequence undergoes cropping and mirroring, the same transformation is applied in the same way to all the frames belonging to the same sequence.

The testing phase is performed considering a video sequence belonging to the first camera as probe and a video sequence belonging to the second camera as gallery. In this phase, we use up to to 128 frames to form a sequence. The frames are always the starting frames for the probe, and the ending frames for the gallery. If this is not possible, because a person's sequence does not have enough frames, we consider all the available frames.

All tests are performed 10 times with different seeds, each time presenting the model different people for training and testing.

\subsection{Results}
First we compared the results of our model when using different numbers of filters for the Spatio-Temporal Attention Module, as shown in Table \ref{tab:filters}. We found that performance increases when generating more filters, but with four or more the model saturates and the performance starts decreasing.

\begin{table}[]
\centering
\caption{Average results obtained using an increasing number of filters.\label{tab:filters}}
    \begin{tabularx}{\textwidth}{ >{\centering\arraybackslash}X|>{\centering\arraybackslash}X|>{\centering\arraybackslash}X|>{\centering\arraybackslash}X|>{\centering\arraybackslash}X }
    \hline
    \multicolumn{5}{c}{Average results using different number of filters} \\
    \hline
    \#filters & Rank-1 & Rank-5 & Rank-10 & Rank-20 \\
    \hline
    {0} & {60.5} & {84.8} & {93} & 96.9 \\
    {1} & {59.4} & {85.7} & {93.2} & 97.4 \\
    {2} & {63} & {\textbf{87.7}} & {93.9} & 97.3 \\
    {3} & {\textbf{63.3}} & {87.4} & {\textbf{94}} & \textbf{97.8} \\
    {4} & {59.6} & {87.2} & {93.9} & 97.7
    \end{tabularx}
\end{table}

Second, we present experimental results with 3 filters on sequences of varying lengths between 2 and up to 128 frames, and the results are shown in Table \ref{tab:results}. Note that if a person's sequence does not have enough frames, we still consider all the available frames and that in all cases the training has been performed using a fixed length sequence of 16. As one could expect, it is confirmed that the performance increases as the number of frames in sequence of frames grows, as also noted by \cite{mclaughlin2016recurrent}. Since the average sequence length in the dataset is 73, the performance does not increase much between 64 and 128, because most sequences are not long enough to benefit from the additional length.
% Note that we trained the model on fixed length sequences of 16, and the performance increase with the length, as already noted by \cite{mclaughlin2016recurrent}.

\begin{table}[]
\centering
\caption{Average results with different sequence lengths (expressed in frames).\label{tab:results}}
\begin{tabularx}{\textwidth}{ >{\centering\arraybackslash}X|>{\centering\arraybackslash}X|>{\centering\arraybackslash}X|>{\centering\arraybackslash}X|>{\centering\arraybackslash}X }
\hline
\multicolumn{5}{c}{Average results with different sequence lengths} \\ \hline
{Length} & {Rank-1} & {Rank-5} & {Rank-10} & Rank-20 \\ \hline
{2} & {16.7} & {37.7} & {50.9} & 64.6 \\
{4} & {22.7} & {46.9} & {60.3} & 72.6 \\
{8} & {31.7} & {59.3} & {71.3} & 84.2 \\
{16} & {43.8} & {72.6} & {83.9} & 91.4 \\
{32} & {53.9} & {80.7} & {89} & 95.3 \\
{64} & {61} & {85.6} & {92.5} & 96.7 \\
{128} & {63.3} & {87.4} & {94} & 97.8
\end{tabularx}
\end{table}

Finally, we present the comparison of our model with the state-of-the-art in Table \ref{tab:compare}. Despite beeing a simple architecture, our solution outperforms other methods proposed in the literature on 2 metrics out of 4. %, improving previous results by 2\% on Rank-1 and 1\% on Rank-5, and only 0.9\% and 0.3\% less on Rank-10 and Rank-20 respectively.
Note that \cite{rao2018videobased} claim better results on their paper, but, in order to provide a fair comparison, we re-ran their provided code on our dataset splits. In addition, for the sake of completeness we report the performance of  \cite{DBLP:journals/corr/LiuYO17} as well, even if their testing protocol is not directly comparable with the others, as they always use all the available frames.

%% punti a favore e punti deboli
%Our approach provides state-of-the art performance while maintaining a simple architecture. 
The simplicity of our architecture comes from the choice of making the spatial and temporal module work jointly. In fact their output is merged in order to, hopefully, get the best of the two and select only the most relevant information obtained by their combination.

%This is, however, one way to use the outputs of the two modules; a different way to merge them could potentially lead to better performances. Moreover, since the last part of the model works solely with the results of the two modules' computations, there's no guarantee that the original information coming directly from the frames is maintained or processed properly until the end.
%Our model it's simple but not lightweight. Training lasts 30 seconds per epoch, over 1500 epoch, due to heaviness of the Attentive-ConvLSTM. It is also uncertain if the attentive module is actually identifying the most relevant part of the person and not some equally representative background noise, or if it's working correctly even with such small input images.

\begin{table}[]
\centering
\caption{Comparison with state-of-the-art methods.\label{tab:compare}}
\begin{tabularx}{\textwidth}{ >{\centering\arraybackslash}c|>{\centering\arraybackslash}X|>{\centering\arraybackslash}X|>{\centering\arraybackslash}X|>{\centering\arraybackslash}X }
\hline
\multicolumn{5}{c}{iLIDS-VID} \\ \hline
{Methods} & {Rank-1} & {Rank-5} & {Rank-10} & Rank-20 \\ \hline
{\textbf{Proposed Approach }} & {\textbf{63.3}} & {\textbf{87.4}} & {94} & 97.8 \\
{Rao et al.\cite{rao2018videobased}} & {62.2\footnote{These results were obtained in our tests on the code provided, and are substantially lower than claimed in the paper}} & {86.8} & {\textbf{94.8}} & 97.8 \\
{Xu et al.\cite{yi2014deep}} & {62} & {86} & {94} & \textbf{98} \\
{Zhang et al.\cite{DBLP:journals/corr/ZhangHL17}} & {60.2} & {85.1} & {-} & 94.2 \\
{McLaughlin et al.\cite{mclaughlin2016recurrent}} & {58} & {84} & {91} & 96 \\
{Zhengl et al.\cite{zheng2016mars}} & {53} & {81.4} & {-} & 95.1 \\
{Yan et al.\cite{yan2016person}} & {49.3} & {76.8} & {85.3} & 90.1 \\ \hline
{Liu et al.\cite{DBLP:journals/corr/LiuYO17}} & {68\footnote{Results are shown for completeness, but are not directly comparable}} & {86.8} & {95.4} & 97.4 \\
% questo tizio non usa il nostro stesso metodo di test, prende sempre la sequenza più lunga mentre noi ci limitiamo a 128
\end{tabularx}

\end{table}

\section{Conclusions}
We described a novel architecture which exploits a single attentive network to extract both spatial and temporal features to perform video-based person Re-Identification, providing state-of-the-art performance on the recent challenging iLIDS-VID dataset. %Our approach improved current Automatic Person Re-Identification state-of-the-art by 2\% on Rank-1 and 1\% on Rank-5.

While the improvement obtained is not groundbreaking, the experiments confirm that employing a joint spatial and temporal attention mechanism can help pushing higher the performances in the field of person Re-Identification using only simple neural networks.

Our experiments confirms that using a longer sequence of frames brings to better performance. Analogously, one may think that using an higher number of filters will always lead to better results; however this is true up to a certain point: our experiments shows that using 3 attentive filters is better than using none, but going above this number leads to a degradation in performance.

Future work will validate the results obtained in this study performing the reported experiments on other datasets. %Regarding the network proposed, the Attentive Convolutional LSTM could be used together with more complex architectures for the extraction of spatial and temporal features. Moreover, since the development of attention models is constantly moving forward, both the spatial and temporal modules proposed in this work can be switched with more complex, and hopefully more performing, architectures that implement the attention mechanism.}

\footnotetext[1]{These results were obtained in our tests on the code provided by the authors, and are substantially lower than claimed in the paper}
\footnotetext[2]{Results are shown for completeness, but are not directly comparable}

\section*{Acknowledgements}
This project was partially supported by the FVG P.O.R. FESR 2014-2020 fund, project \lq\lq Design of a Digital Assistant based on machine learning and natural language, and by the ``PREscriptive Situational awareness for cooperative autoorganizing aerial sensor NETworks'' project CIG68827500FB.
\rq\rq.

\bibliographystyle{ieeetr}
\bibliography{mybibliography}
\end{document}